\pdfoutput=1
\documentclass[letterpaper]{article}
\usepackage{aaai}
\usepackage{times}
\usepackage{helvet}
\usepackage{courier}

\usepackage{graphicx} 
\usepackage{subfigure}
\usepackage{algorithm}
\usepackage{algorithmic}

\usepackage{color}
\usepackage{amsmath}
\usepackage{wrapfig}

\usepackage{bm}
\usepackage{dsfont} 
\usepackage{amssymb} 

\ifdefined\theorem
\else
    \newtheorem{theorem}{Theorem}
\fi
\ifdefined\lemma
\else
    \newtheorem{lemma}[theorem]{Lemma}
\fi
\ifdefined\corollary
\else
    \newtheorem{corollary}[theorem]{Corollary}
\fi

\newcommand{\QED}{\hfill$\;\;\;\rule[0.1mm]{2mm}{2mm}$\\}

\newfont{\msym}{msbm10}

\newcommand{\sign}{{\rm sign}}
\newcommand{\paren}[1]{\left({#1}\right)}
\newcommand{\brackets}[1]{\left[{#1}\right]}
\newcommand{\braces}[1]{\left\{{#1}\right\}}

\newcommand{\abs}[1]{\left\vert{#1}\right\vert}

\newcommand{\comdots}{, \ldots ,}

\newcommand{\argmin}[1]{\underset{#1}{\mathrm{argmin}} \:}
\newcommand{\argmax}[1]{\underset{#1}{\mathrm{argmax}} \:}
\newcommand{\normt}[1]{\left\Vert {#1} \right\Vert^2}

\newcommand{\ind}[1]{\mathds{1}\braces{#1}} 

\newcommand{\beq}[1]{\begin{equation}\label{#1}}
\newcommand{\eeq}{\end{equation}}
\newcommand{\beqa}{\begin{eqnarray}}
\newcommand{\eeqa}{\end{eqnarray}}
\renewcommand{\eqref}[1]{Eq.~(\ref{#1})}
\newcommand{\secref}[1]{Sec.~\ref{#1}}
\providecommand{\algref}[1]{Alg.~\ref{#1}} 
\newcommand{\figref}[1]{Fig.~\ref{#1}}

\newcommand{\thmref}[1]{Theorem~\ref{#1}}

\newcommand{\lemref}[1]{Lemma~\ref{#1}}

\newcommand{\mb}[1]{{\boldsymbol{#1}}}

\newcommand{\vx}{\mathbf{x}}
\newcommand{\vxi}[1]{\vx_{#1}}
\newcommand{\vxii}{\vxi{i}}
\newcommand{\bvx}{\bar{\vx}}
\newcommand{\bvxi}[1]{\bvx_{#1}}

\newcommand{\vzi}[1]{\vz_{#1}}
\newcommand{\vzii}{\vzi{i}}

\newcommand{\vmu}{\mb{\mu}}

\newcommand{\vmui}[1]{\vmu_{#1}}
\newcommand{\vmuii}{\vmui{i}}

\newcommand{\vv}{\mb{v}}

\newcommand{\hvx}{\hat{\vx}}
\newcommand{\hvxi}[1]{\hvx_{#1}}
\newcommand{\hvxii}{\hvxi{i}}

\newcommand{\newstufffroma}[1]{}

\newcommand{\newstufffrom}[1]{}

\newcommand{\oldnote}[2]{}

\newcommand{\commentout}[1]{}

\newcommand{\vz}{\mb{z}}

\usepackage{amssymb} 
\usepackage{textcomp} 

\newcommand{\bbraces}[1]{\Bigg\{#1\Bigg\}}
\newcommand{\expec}[1]{\mathbb{E}\brackets{#1}}
\newcommand{\prob}[1]{\mathbb{P}\paren{#1}}

\newcommand{\vnorm}[1]{\left\Vert#1\right\Vert}

\newcommand{\textapprox}{\raisebox{0.2ex}{\texttildelow}}

\newcommand{\negspaces}{\!\!\!}
\newcommand{\negspace}{\!\!\!\!\!\!}

\renewcommand{\algref}[1]{Alg.~\ref{#1}}

\title{Outlier-Robust Convex Segmentation}
\author{
Itamar Katz \and Koby Crammer\\
Department of Electrical Engineering\\
The Technion -- Israel Institute of Technology\\
Haifa, 32000 Israel\\
\texttt{(itamark@tx,koby@ee).technion.ac.il}
}

\begin{document}

\maketitle
\begin{abstract}
\begin{quote}
We derive a convex optimization problem for the task of segmenting sequential data, which
explicitly treats presence of outliers.
We describe two algorithms for solving this
problem, one exact and one a top-down novel approach,
and we derive a consistency results for the case of two segments and no outliers.
Robustness to outliers is evaluated on
two real-world tasks related to speech segmentation.
Our algorithms outperform baseline segmentation algorithms.
\end{quote}
\end{abstract}

\section{Introduction}\label{sec:intro}
Segmentation of sequential data, also known as change-point detection, is a fundamental problem in the field
of unsupervised learning, and has applications in diverse fields such
as speech processing \cite{brent1999speech,phonemeSeg:2008,Shriberg2000127},
text processing \cite{beeferman1999statistical}, bioinformatics \cite{olshen2004circular}
and network anomaly detection \cite{levy2009detection}, to name a few.
We are interested in formulating the segmentation task as a convex optimization
problem that avoids issues such as local-minima or sensitivity to
initializations. In addition, we want to explicitly incorporate robustness to outliers.
Given a sequence of samples $\{\vxii\}_{i=1}^n$, for
$\vxii\in\mathbb{R}^d$, our goal is to segment it into a few
subsequences, where each subsequence is homogeneous under some criterion. Our starting point is a convex objective
that minimizes the sum of squared distances of samples $\vxii$ from each sample's associated `centroid`, $\vmuii$. Identical adjacent $\vmuii$s identify their corresponding samples as belonging to the same segment.
In addition, some of the samples are allowed to be identified as outliers, allowing reduced loss on these samples. Two regularization terms are added to the objective, in order to constrain the number of detected segments and outliers, respectively. 

We propose two algorithms
based on this formulation, both alternate between detecting outliers, which is solved analytically, and solving the problem
with modified samples 
, which can be solved iteratively.
The first algorithm, denoted by Outlier-Robust Convex Sequential (ORCS)
segmentation, solves the optimization problem exactly, while the
second is a top-down hierarchical version of the algorithm, called
TD-ORCS. We also derive a weighted version of this algorithm, denoted by WTD-ORCS. We show that for the case of $K=2$ segments and no outliers, a specific choice of the weights leads to a solution which recovers the exact solution of an un-relaxed optimization problem.

We evaluate the
performance of the proposed algorithms on 
two speech segmentation tasks,
for both clean sources and sources
contaminated with added non-stationary noise. 
Our algorithms outperform other algorithms in both the clean and outlier-contaminated setting.
Finally, based on the empirical results, we propose a heuristic approach for approximating the number of outliers.

{\bf Notation}: The samples to be segmented are
denoted by $\{\vxii\in\mathbb{R}^d\}_{i=1}^n$, and their associated
quantities (both variables and solutions) are
$\vmuii,\vzii\in\mathbb{R}^d$. The same notation with no subscript,
$\vmu$, denotes the collection of all $\vmuii$'s.
The same holds for $\vx,\vz$. We abuse notation and
use $\vmuii$ to refer to both the `centroid' vector of a segment (these
are not center of mass, due to the regularization term), and to the
indexes of measurements assigned to that segment.

\section{Outlier-Robust Convex Segmentation}\label{sec:outlierRobustObj}
Segmentation is the task of dividing a sequence of $n$ data samples $\{\vxii\}_{i=1}^n$, into $K$ groups of consecutive samples, or segments, such that each group is homogeneous with respect to some criterion. A common choice of such a criterion often involves minimizing the squared Euclidean distance of a sample to some representative sample $\vmuii$. 
This criterion is highly sensitive to outliers and indeed, as we show empirically below, the performance of segmentation algorithms degrades drastically when the data is contaminated with outliers.
It is therefore desirable to incorporate robustness to outliers into the model. We achieve this by allowing some of the input
samples $\vxii$ to be identified as outliers, in which case we do not require $\vmuii$ to be close to these samples.
To this end we propose to minimize:
\begin{align*}
    \min_{\vmu,\vz}\bbraces{&\frac{1}{2}\sum_{i=1}^n\normt{\vxii-\vzii-\vmuii}} \\
    \textrm{ s.t. } & \sum_{i=1}^{n-1}\ind{\vnorm{\vmui{i+1}-\vmuii}_p>0}\leq K-1~,\\ 
                    & \sum_{i=1}^n\ind{\vnorm{\vzii}_q>0}\leq M~,
\end{align*}
where $p,q\geq 1$. Considering samples $\vxii$ for which $\vzii=0$, the objective measures the loss of replacing a point $\vxii$ with some shared point $\vmuii$, and can be thought of as minus the log-likelihood under Gaussian noise. Samples $i$ with $\vzii\neq0$ are intuitively identified as outliers. The first constraint bounds the number of segments by $K$,
while the second constraint bounds the number of outliers by $M$.
The optimal value for a nonzero $\vzii$ is to set $\vzii=\vxii-\vmuii$,
making the contribution to the objective zero, and thus in practice
ignoring this sample, treating it as an outlier. We note that a similar approach to robustness was employed by \cite{forero2011outlier} in the context of robust clustering, and by \cite{mateos2012robust} in the context of robust PCA.
Since the $\ell_0$ constraints results in a non convex
problem, we use a common practice and replace it with a convex surrogate $\ell_1$ norm which induces
sparsity. For the $\vmuii$ variables it means that for most samples we will have $\vmui{i+1}-\vmui{i}=0$, allowing the identification of the corresponding samples as belonging to the same segment.
For the $\vzii$ variables, it means that most will satisfy $\vzii=0$ and for some of them, the
outliers, otherwise. We now incorporate the relaxed constraints into the objective, and in addition consider a slightly more general formulation in which we allow weighting of the summands in the first constraint. We get the following optimization problem:
\begin{align}\label{eq:outlierRobustObjective}
    \min_{\vmu,\vz}\bbraces{ &\frac{1}{2}\sum_{i=1}^n\normt{\vxii-\vzii-\vmuii}+\lambda\sum_{i=1}^{n-1}w_i\vnorm{\vmui{i+1}-\vmuii}_p\\
    &+\gamma\sum_{i=1}^n\vnorm{\vzii}_q}~,\nonumber
\end{align}
where $w_i$ are weights to be determined.
The parameter $\lambda>0$ can be thought of as a tradeoff parameter between the first term which is minimized with $n$ segments, and the second term which is minimized with a single segment. As $\lambda$ is decreased, it crosses values at which there is a transition from $K$ segments to $K+1$ segments, in a phase-transition like manner where $1/\lambda$ is the analog of temperature. The parameter $\gamma>0$ controls the amount of
outliers, where for $\gamma=\infty$ we enforce $\vzii=0$ for all samples,
and for $\gamma=0$ the objective is optimal for $\vzii=\vxii-\vmuii$, and thus all
samples are in-fact outliers.
Alternatively, one can think of $\lambda,\gamma$ as the Lagrange multipliers of a constrained optimization problem. In what follows we consider $p=2$ and $q=1,2$, focusing empirically on $q=2$. Note that $q=1$ encourages sparsity of coordinates of $\vzii$, and not of the vector as a whole.
This amounts to outliers being modeled as noise 
in few features or samples, respectively.

\subsection{Algorithms}\label{sec:algorithms}
The decoupling between $\vmu$ and $\vz$ allows us to optimize
\eqref{eq:outlierRobustObjective} in an alternating manner, and we call this algorithm 
Outlier-Robust Convex Sequential (ORCS) segmentation.
Holding $\vmu$ constant, optimizing
over $\vz$ is done analytically by noting that
\eqref{eq:outlierRobustObjective} becomes the definition of the
proximal operator evaluated at $\vxii-\vmuii$, for which a closed-form solution exists.
For $q=1$ the objective as a function of $\vz$
is separable both over coordinates and over data samples, and the proximal
operator is the shrinkage-and-threshold operator evaluated at each
coordinate $k$:
\begin{equation*}
    \mathrm{prox}_\gamma\paren{v_k}=\sign\paren{v_k}\cdot\max\braces{0,\abs{v_k}-\gamma}.
\end{equation*}
However, we are interested in zeroing some of the $\vzii$'s as a whole, so we set $q=2$. In this case, the objective is separable over data samples, and the proximal operator is calculated to be:
\begin{align}\label{eq:proxOp_l2}
\mathrm{prox}_{\gamma}\paren{\vv} = \vv\cdot\max\braces{0,1-\frac{\gamma}{\vnorm{\vv}_2}}.
\end{align}
Holding $\vz$ constant, optimizing over $\vmu$ is done by defining
$\hvxii\triangleq \vxii-\vzii$, which results in the following optimization problem:
\begin{equation}\label{eq:segmentationObjective_mu}
 \min_\vmu\sum_{i=1}^n\normt{\hvxii-\vmuii}  + \lambda\sum_{i=1}^{n-1}w_i\vnorm{\vmui{i+1}-\vmuii}_2.
\end{equation}
Note that \eqref{eq:segmentationObjective_mu} is equivalent to \eqref{eq:outlierRobustObjective} with no outliers present. We also note that if we plug the analytical solution for the $\vzii$s into \eqref{eq:segmentationObjective_mu} (via the $\hvxii$s), the loss term turns out to be the multidimensional equivalent of the Huber loss of robust regression.
We now discuss two approaches for solving \eqref{eq:segmentationObjective_mu}, either exactly or approximately.

\paragraph{Exact solution of \eqref{eq:segmentationObjective_mu}:}
The common proximal-gradient approach \cite{bach2011convex,Beck_FISTA:2009} for solving non-smooth convex problems has in this case the disadvantage of convergence time which grows linearly with the number of samples $n$. The reason is that the Lipschitz constant of the gradient of the first term in \eqref{eq:segmentationObjective_mu} grows linearly with $n$, which results in a decreasing step size. An alternative approach is to derive the dual optimization problem to \eqref{eq:segmentationObjective_mu}, analogously to the derivation of \cite{Beck_TV:2009} 
in the context of image denoising. The resulting objective is smooth and has a bounded Lipschitz constant independent of $n$.
Yet another approach was proposed by \cite{bleakley2011group} for the task of change-point detection, who showed that under a suitable change of variables \eqref{eq:segmentationObjective_mu} can be formulated as a group-LASSO regression \cite{tibshirani1996regression,yuan2006model}.
\paragraph{Approximate solution of \eqref{eq:segmentationObjective_mu}:}\label{par:seg_approx_solution}
Two reasons suggest that deriving an alternative algorithm for solving \eqref{eq:segmentationObjective_mu} might have an advantage. 
First, the parameter $\lambda$
does not allow direct control of the resulting number of segments, and in many use-cases such a control is a desired property.
Second, as mentioned above,
\cite{bleakley2011group} showed that \eqref{eq:segmentationObjective_mu} is equivalent to group-LASSO regression, under a suitable change of variables. It is known from the theory of LASSO regression
that certain conditions on the design matrix 
must hold in order for perfect detection of segment boundaries to be possible. Unfortunately, these conditions are violated for the objective in \eqref{eq:segmentationObjective_mu} 
; see \cite{levy2007catching} and references therein. Therefore a non-exact solution has a potential of performing better, at least in some situations. We indeed encountered this phenomenon empirically, as is demonstrated in \secref{sec:experimentalResults}.
Therefore we also derive an alternative top-down, greedy algorithm, which finds a segmentation into $K$
segments, where $K$ is a user-controlled parameter. The algorithm works in rounds. On each round it picks a segment of a
current segmentation, and finds the optimal segmentation of it into two subsequences.
We start with the following lemma, which gives an analytical rule which solves \eqref{eq:segmentationObjective_mu} for the case of $K=2$ segments.
\begin{lemma}\label{lem:analyticalSolutionK=2}
Consider the optimal solution of \eqref{eq:segmentationObjective_mu} for the largest parameter $\lambda$ for which there are $K=2$ segments, and denote this value of the parameter by $\lambda^*$. Denote by $i^*$ the associated splitting point into $2$ segments, i.e. samples $\hvxii$ with $i\leq i^*$ belong to the first segment, and otherwise belong to the second segment. Then 
$i^*\paren{\vx} =\argmax{1\leq i\leq n-1}g\paren{i,\vx}$, where:
\begin{align}\label{eq:lambdaStarWeighted}
    g\paren{i,\vx} = \bbraces{\frac{i(n-i)}{w_in}\vnorm{\bvxi{2}(i)-\bvxi{1}(i)}_2},
\end{align}
and $\bvxi{1,2}(i)$ are the means of the first and second segments, respectively, given that the split occurs after the $i$th sample. In addition, $\lambda^*\paren{\vx} = g\paren{i^*,\vx}$.
\end{lemma}
The proof is given in 
the supplementary material.
This result motivates a top-down hierarchical segmentation algorithm, which chooses at each iteration to split the segment
which results in the maximal decrease of the sum-of-squared-errors criterion. Note that we
cannot use the criterion of minimal increment to the objective in \eqref{eq:segmentationObjective_mu}, since by continuity of
the solution path, there is no change in the objective at the splitting from $K=1$ to $K=2$ segments. The top-down algorithm can be implemented in $\mathcal{O}\paren{nK}$.
It has the advantage that no search in the solution path is needed in case $K$ is known, and
that this search can be made efficiently in case where $K$ is not known. The top-down approach is used in the algorithm presented in \secref{sec:robust_top_down}.

From the functional form of $g\paren{i,\vx}$ in \eqref{eq:lambdaStarWeighted} it is clear that in the unweighted case ($w_i=1$ for all $i$), the solution is biased towards segments of approximately the same length, because of the $i\paren{n-i}$ factor.
We now show that a specific choice of $w_i$ exactly recovers the solution to the unrelaxed optimization problem, where the regularization term in \eqref{eq:segmentationObjective_mu} is replaced with the $\ell_0$ constraint, that is $\sum_{i=1}^{n-1}\ind{\vnorm{\vmui{i+1}-\vmuii}_2>0}=1$. This is formulated by the following lemma:
\begin{lemma}\label{lem:weighted_solution_l0_l1}
    Consider the case of two segments, $K=2$, and denote by $n_w^*$ the minimizer of \eqref{eq:lambdaStarWeighted} with $w_i=\sqrt{i\paren{n-i}}$. Then the split into two segments found by solving the following: 
    \begin{align}\label{eq:segmentationObjective_unrelaxed_K=2}
        \argmin{\vmu}&\braces{\sum_{i=1}^n\normt{\vxii-\vmuii}},\\
        \mathrm{s.t.}\ &\sum_{i=1}^{n-1}\ind{\vnorm{\vmui{i+1}-\vmuii}_2>0}=1~,\nonumber
    \end{align}
    is also given by $n_w^*$.
\end{lemma}
The proof appears in 
the supplementary material.
We note that the same choice for $w_i$ was derived by \cite{bleakley2011group} from different considerations based on a specific noise model for the stochastic process generating the data. In this sense our derivation is more general, as it does not make any assumptions about the data.

\subsubsection{Robust top-down algorithm}\label{sec:robust_top_down}
We now propose a robust top-down algorithm for approximately optimizing \eqref{eq:outlierRobustObjective}.
For a fixed value of $\vmu$, using \eqref{eq:proxOp_l2} we can
calculate analytically which of the $\vzii$s represent a detected
outlier. These are $\vzii$s which satisfy
$\vnorm{\vzii}_2>\gamma$. This allows us to calculate the value
$\gamma^*$ for which the first outlier is detected as having a
non-zero norm. Furthermore, for $\lambda=\lambda^*,\ \gamma=\gamma^*$ we know that $\vzii=0$ for all $i=1\comdots n$, and therefore
$\vmuii=\mathrm{mean}\paren{\vx-\vz}=\bvx$ for all $i=1\comdots n$, and we
can find $\gamma^*$ analytically:
\begin{equation}\label{eq:gmaStar}
    \gamma^*=\max_i\braces{\vnorm{\vxii-\bvx}_2}~,
\end{equation}
where the index $i^*$ at which the maximum is attained is the index to the first detected outlier. The value of $\lambda^*$ is found as given in \lemref{lem:analyticalSolutionK=2}, with the replacement of each $\vxii$ with $\hvxii$ as defined above. We note that the values $\lambda^*,\ \gamma^*$ are helpful for finding a solution path, since they allow to exclude parameters which result in trivial solutions.

In the case where $\lambda=\lambda^*$ we can extend \eqref{eq:gmaStar} for any number $M>1$ of outliers, by simply looking for the first $M$ vectors $\vxii-\bvx$ with the largest norm.
In this case it no longer holds true that $\vzii=0$ for all $i=1\comdots n$, so
we have to use the alternating optimization in order to find a
solution. However, each iteration is now solved analytically and
convergence is fast compared to the case $\lambda<\lambda^*$ where we
do not have an analytical solution for the optimization over $\vmu$. This
result motivates the top-down version of the ORCS algorithm. We
denote the algorithm by TD-ORCS for the unweighted case ($w_i=1$, $i=1\comdots n$), and by WTD-ORCS when using the weights given in \lemref{lem:weighted_solution_l0_l1}. The number of required segments $K$ and number of
required outliers $M$ is set by the user. In
each iteration the algorithm chooses the segment-split which results in
the maximal decrease in the squared loss. Whenever a segment is split,
the number of outliers belonging to each sub-segment is kept and used
in the next iteration, so the overall number of outliers equals $M$ at
all iterations. The algorithm is summarized in \algref{alg:topdownORCS}.

\begin{algorithm}[!t]
\footnotesize
 \caption{Top-down outlier-robust hierarchical segmentation}\label{alg:topdownORCS}
  \begin{algorithmic}
    \STATE \textbf{Input:} Data samples $\braces{\vxii}_{i=1}^n$.
    \STATE \textbf{Parameters:} Number of required segments $K$, number of required outliers $M$, weights $\braces{w_i}_{i=1}^{n-1}$.
    \STATE \textbf{Initialize:} Segments boundaries $\mathrm{B}=\braces{1,n}$, current number of segments $k=\abs{B}-1$, number of outliers for each segment $M_j=M$ (for $j=1$).
    \WHILE{$k<K$}
        \FOR{$j=1,...,k$}
            \STATE  Set $S_j=\braces{\vxi{B_j},..,\vxi{B_{j+1}-1}}$.
            \WHILE {not converged}
                \STATE Split segment $S_j$ into sub-sequences $S_{1,2}$, using the solution to
                \eqref{eq:lambdaStarWeighted} at $\vx-\vz$, with weights $\braces{w_i}$.
                \STATE Find $\gamma$ for $M_j$ outliers, using the extension of \eqref{eq:gmaStar} to $M_j$ outliers.
                \STATE Set $\vzii=\mathrm{prox}_{\gamma}\paren{\vxii-\bvx}$, for $i=1\comdots n$.
            \ENDWHILE
            \STATE Calculate the mean $\bvxi{j}$ of segment $S_j$, and the means of $S_{1,2}$, denoted by $\bvxi{1}$ and $\bvxi{2}$.
            \STATE Set \(L(j)=
              \sum \limits_{i\in S_1}\!\normt{\vxii-\bvxi{1}}
              \!\!+\!\!
              \sum\limits_{i\in S_2}\!\normt{\vxii-\bvxi{2}}
              \!\!-\!\!
              \sum \limits_{i\in S_j}\!\normt{\vxii-\bvxi{j}}\).
        \ENDFOR
        \STATE Choose segment $j^* = \arg\max_j  L(j)$ and calculate the splitting
        point $n_{j^*}$
        \STATE Add the new boundary $\vxi{B_{j^*}} + n_{j^*} + 1$ to the (sorted) boundary list $B$.
        \STATE Set $M_j$ and $M_{j+1}$ to the number of $\vzii$s with non-zero norm in segment $S_j$ and $S_{j+1}$, respectively.
    \ENDWHILE
\end{algorithmic}
\end{algorithm}

\subsection{Analysis of \lemref{lem:analyticalSolutionK=2} for K=2}\label{sec:detectionBounds}
We now bound the probability that the solution as given by \lemref{lem:analyticalSolutionK=2} fails to detect the correct boundary. We use the derived bound to show that the weights given in \lemref{lem:weighted_solution_l0_l1} are optimal in a sense explained below.
For simplicity we analyze the one dimensional case $\vxii\in\mathbb{R}$, and we show later how the results generalize to multidimensional data.

We assume now that the data sequence is composed of two subsequences
of lengths $n_1$ and $n_2$, each composed of samples taken iid
from some probability distributions with means $\mu_1$
and $\mu_2$ respectively, and define $\Delta\mu\triangleq\mu_2-\mu_1$. We further
assume that the samples are bounded, i.e. $\abs{\vxii}\leq M/2,\;i=1\comdots n$,
for some positive constant $M$.
We set $w_i=1$ for all $i=1\comdots n-1$, and quote results for the weighted case where relevant. We note that $n_1$ and $n_2$ represent the
ground-truth, and not a variable we have to optimize.
We parameterize the sample-index argument of $g\paren{\cdot}$ in \eqref{eq:lambdaStarWeighted} as $i=n_1+m$ (and similarly $i^*=n_1+m^*$), that is we measure it
relatively to the true
splitting point $n_1$. For ease of notation, in what follows we substitute $g\paren{m}$ for $g\paren{n_1+m}$.
Without loss of generality, we treat the case where
$m\geq0$. Note that $m^*\neq0$ if $g(0)<g(m)$ for some $m>0$.
The probability of this event is bounded:
\begin{align}\label{eq:1D_errBound_m}
    \prob{g(0)<g(m)} 
    \leq 2\exp\paren{-Cm},
\end{align}
for $C=\paren{2\Delta\mu^2n_1^2}/\paren{M^2n^2}$.
The proof is given in 
the supplementary material.

Note that in order for the bound to be useful, the true segments should not be
too long or too short, in agreement with the motivation for using weights given
before \lemref{lem:weighted_solution_l0_l1}.
We now use \eqref{eq:1D_errBound_m} to prove the following theorem:
\begin{theorem}\label{thm:1D_errBound}
    Consider a sequence of $n$ variables as described
    above. Given $\delta\in(0,1)$ set
    $m_0=\frac{\log\paren{2n_2/\delta}}{C}$. Then, the probability that
    the solution 
    $i^*=n_1+m^*$ as given in \lemref{lem:analyticalSolutionK=2} 
    is no less than
    $m_0$ samples away from the true boundary is bounded,
    $\mathbb{P}\paren{m^*\geq m_0}\leq\delta$~.
\end{theorem}
The proof appears in 
the supplementary material.

\begin{figure}[!t]
        \begin{minipage}{.47\textwidth}
            \newcommand{\factor}{.49} 
            \centering
            \subfigure[$n_1=20$\label{fig:weighted_bounds2}]{\includegraphics[width=\factor\linewidth]{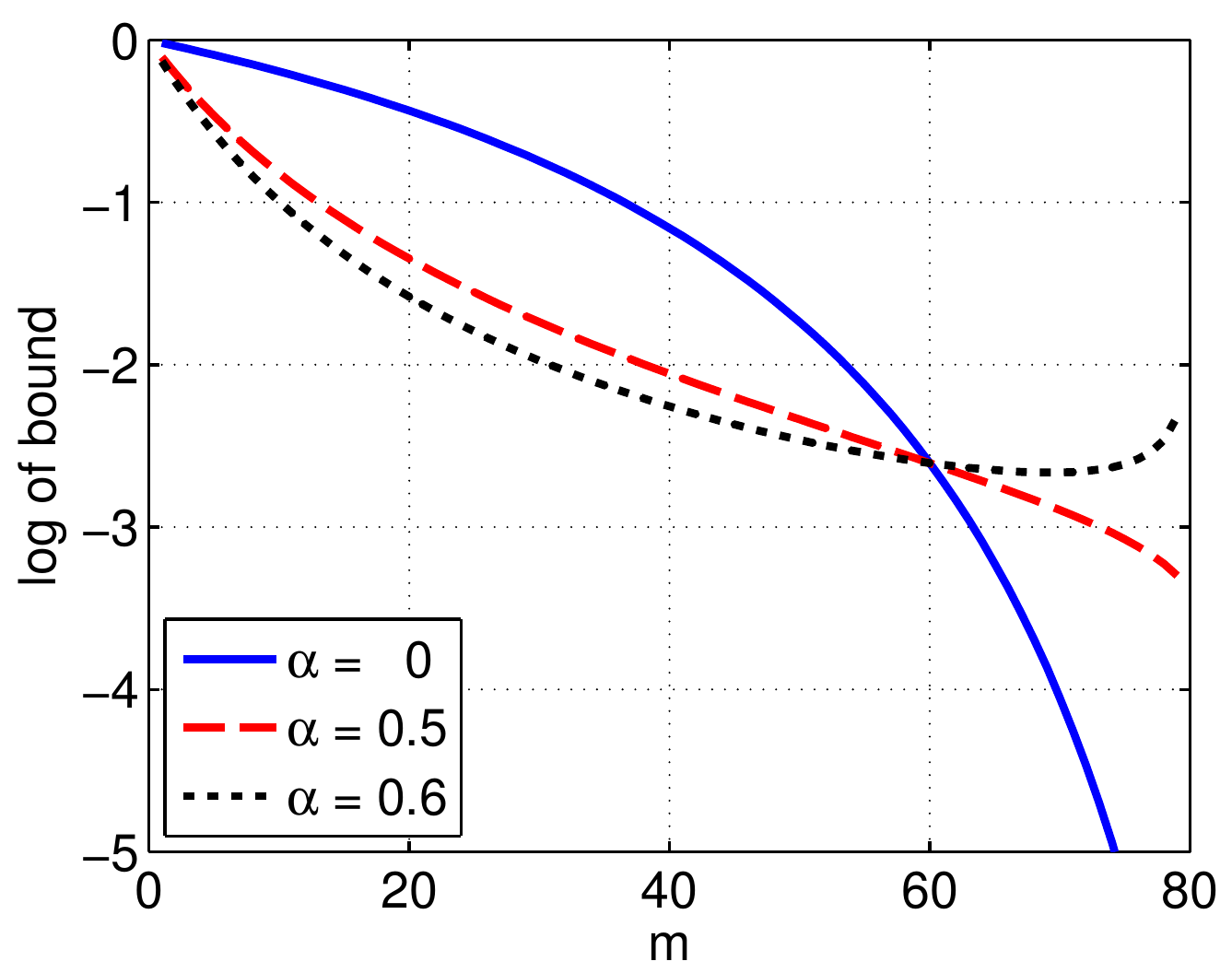}}
            \subfigure[$n_1=40$\label{fig:weighted_bounds4}]{\includegraphics[width=\factor\linewidth]{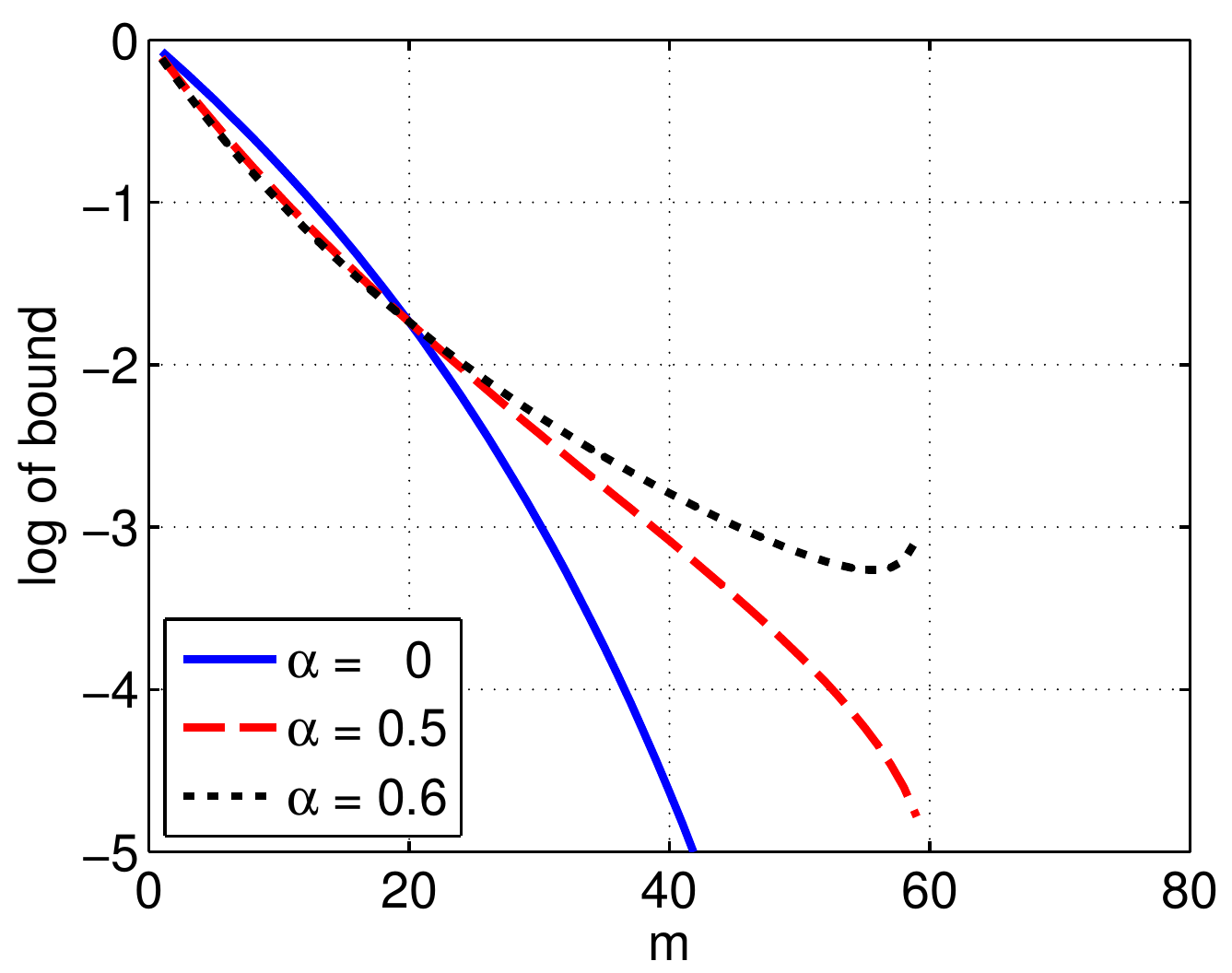}}
            \caption{Bounds on 
            $\prob{g(0)<g(m)}$
            as a function of the distance $m$ from the true boundary $n_1$, for $n=100$, two values of $n_1$, and various values of the weighting parameter $\alpha$. The case of $\alpha=0$ amounts to uniform weights.}\label{fig:weighted_bounds}
            \quad
            \renewcommand{\factor}{.6}
            \centering
            \includegraphics[width=\factor\linewidth]{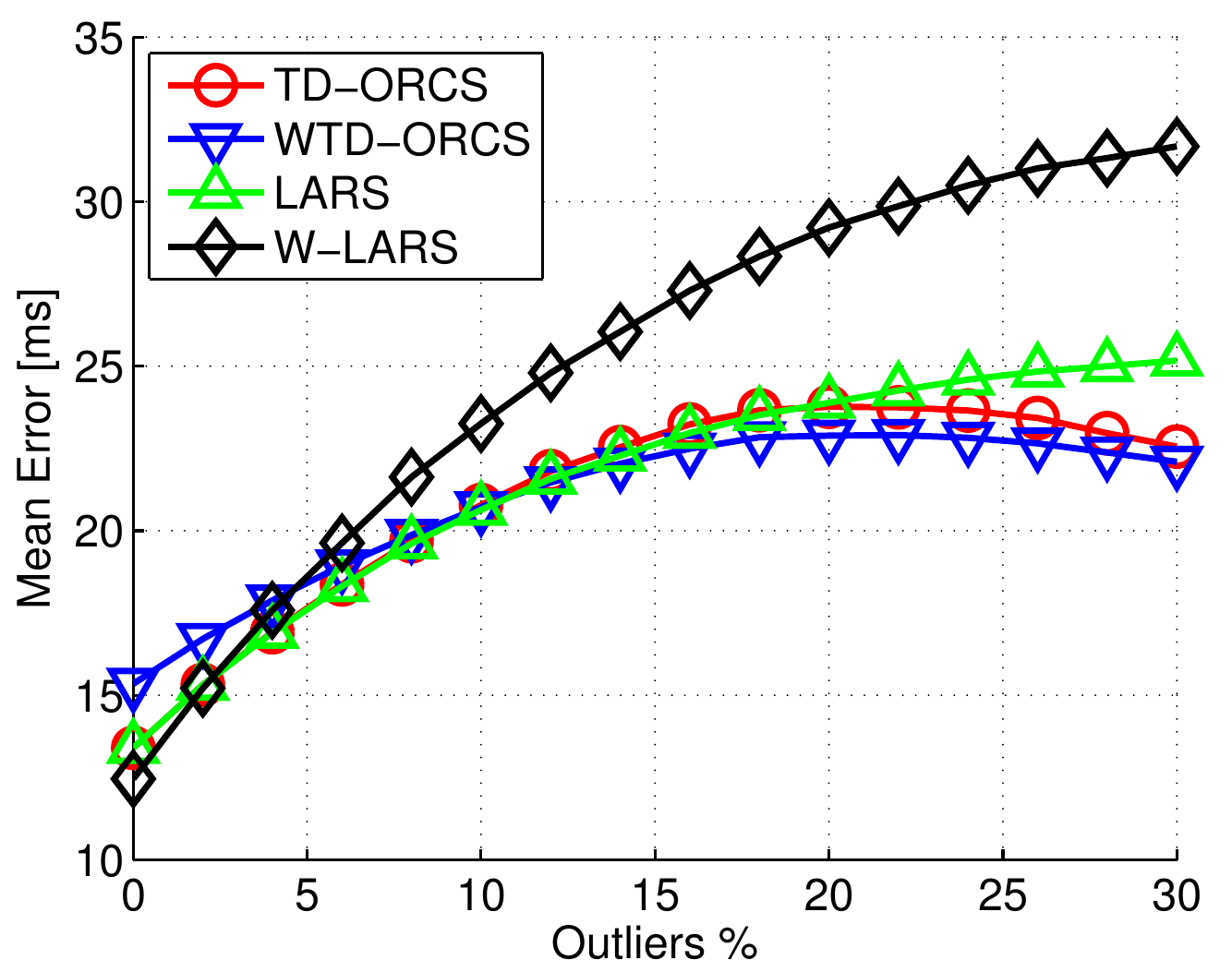}
            \caption{Mean error (absolute time difference to ground-truth boundary)
            of four algorithms on TIMIT phonetic data contaminated with outliers. See text for a list of algorithms compared.}\label{fig:biphonesSeg_robust}
        \end{minipage}
\end{figure}

Considering the weighted case with arbitrary $w_i$, we repeated the calculation for the bound on
$\prob{g(0)<g(m)}$.
To illustrate the influence of the weights on the bound, we heuristically parameterize $w_j=\paren{j\paren{n-j}}^\alpha$ for some $\alpha\in\brackets{0,1}$\footnote{This parametrization is motivated by \eqref{eq:lambdaStarWeighted}}.
The bound as a function of $m$ is illustrated in \figref{fig:weighted_bounds} for several values of $\alpha$.
It is evident that indeed $\alpha>0$ achieves a faster decaying bound for small $n_1$, and that $\alpha=0.5$ is optimal in the sense that for $\alpha>0.5$ the bound is no longer a monotonous function of $m$. This agrees with the weights given by \lemref{lem:weighted_solution_l0_l1}. We note that this specific choice of the weights was derived as well by \cite{bleakley2011group} by assuming a Gaussian noise model, while our derivation is more general.
%
Finally, we note that generalizing the results for multidimensional
data is done by using the fact that for any two vectors
$a,b\in\mathbb{R}^d$, it holds true that
$\prob{\vnorm{a}_2\leq\vnorm{b}_2}\leq\sum_i\prob{\abs{a_i}\leq\abs{b_i}}$. Thus
generalizing
\eqref{eq:1D_errBound_m}
for $d>1$ is
straightforward. While the bound derived in this way will have a
multiplicative factor of the dimension of the data $d$, it is still
exponential in the number of samples $n$.

\section{Empirical Study}\label{sec:experimentalResults}
We compared the unweighted (\mbox{TD-ORCS}) and weighted
(\mbox{WTD-ORCS}) versions of our top-down algorithm to LASSO and group fused \mbox{LARS} \footnote{http://cbio.ensmp.fr/~jvert/svn/GFLseg/html/} of \cite{bleakley2011group},
which are based on reformulating \eqref{eq:segmentationObjective_mu} as group
LASSO regression, and solving the optimization problem either exactly or approximately.
Both the \mbox{TD-ORCS} and \mbox{LARS} algorithms
have complexity of $\mathcal{O}\paren{nK}$.
We also report results for a Bayesian change-point detection algorithm \mbox{(BCP)}, as formulated by \cite{erdman2008fast}. We note that we experimented with a left-to-right Hidden Markov Model (HMM) for segmentation. We do not report results for this model, as its performance was inferior to the other baselines.


\subsection{Biphones subsequences segmentation}\label{sec:biphonesSegExperiment}
In this experiment we used the \mbox{TIMIT} corpus 
\cite{garofolo1988timit}. Data include $4,620$ utterances with annotated phoneme boundaries, amounting to
more than $170,000$ boundaries. Audio was divided into frames
of $16$ms duration and
$1$ ms hop-length, each represented with $13$
\mbox{MFCC} coefficients. The task is to find the boundary between two
consecutive phonemes (biphones), and performance is evaluated as the mean
absolute distance between the detected and
ground-truth boundaries.
Since the number of segments is $K=2$ the \mbox{ORCS} and the
\mbox{TD-ORCS} algorithms are essentially identical, and the same holds for \mbox{LASSO} and \mbox{LARS}.
Outliers were incorporated by adding short ($0.25$ frame-length) synthetic transients to the audio source.
The percentage
of outliers reflects the percentage of contaminated frames. Results are shown in \figref{fig:biphonesSeg_robust} as the mean error as a function of outlier percentage.
For low fraction of outliers, all algorithms perform the same, except
\mbox{WTD-ORCS}, which is slightly worse. For about $15\%$ outliers, the
performance of \mbox{W-LARS} degrades to \textapprox$27$ms mean error vs
\textapprox$22$ms for the rest. For $30\%$ outliers
both \mbox{TD-ORCS} algorithms outperform all other algorithms.
The counter-intuitive drop of error at high outliers rate for the \mbox{TD-ORCS} algorithms
might be the result of over-estimating the number of outliers. We plan to further investigate this phenomenon in future work.

We also compared our algorithm to RD, which \cite{phonemeSeg:2008} found to be the best
among five different objective functions, and was not designed for
treating outliers. In this setting (no outliers) the RD algorithm achieved
$15.1$ms mean error, while TD-ORCS achieved $13.4$ms, with $95$\% confidence interval (not
reported for the RD algorithm) of $0.1$.


\subsection{Radio show segmentation}
In this experiment we used a $35$ minutes, hand-annotated audio
recording of a radio talk show, composed of different
sections such as opening title, monologues, dialogs, and songs.
A detected segment boundary is considered a true positive if it falls within a tolerance window of two frames around a ground-truth boundary. Segmentation quality is commonly measured using the F measure, which is defined as $2pr/\paren{p+r}$, where $p$ is the precision and $r$ is the recall.
Instead, we used the R measure introduced by \cite{Rasanen2009}, which is more robust
to over-segmentation. It is defined as $R\triangleq1-0.5\paren{\abs{s_1}+\abs{s_2}}$, where $s_1\triangleq\sqrt{\paren{1-r}^2+\paren{r/p-1}^2}$ and $s_2\triangleq\paren{r-r/p}/\sqrt{2}$.
The R measure satisfies $R\leq1$,
and $R=1$ only if $p=r=1$.

\begin{figure}[!t]
        \newcommand{\factor}{.45}
        \begin{minipage}{.47\textwidth}
            \centering
            \subfigure[Likelihood matrix\label{fig:likemat}]{\includegraphics[width=.44\linewidth]{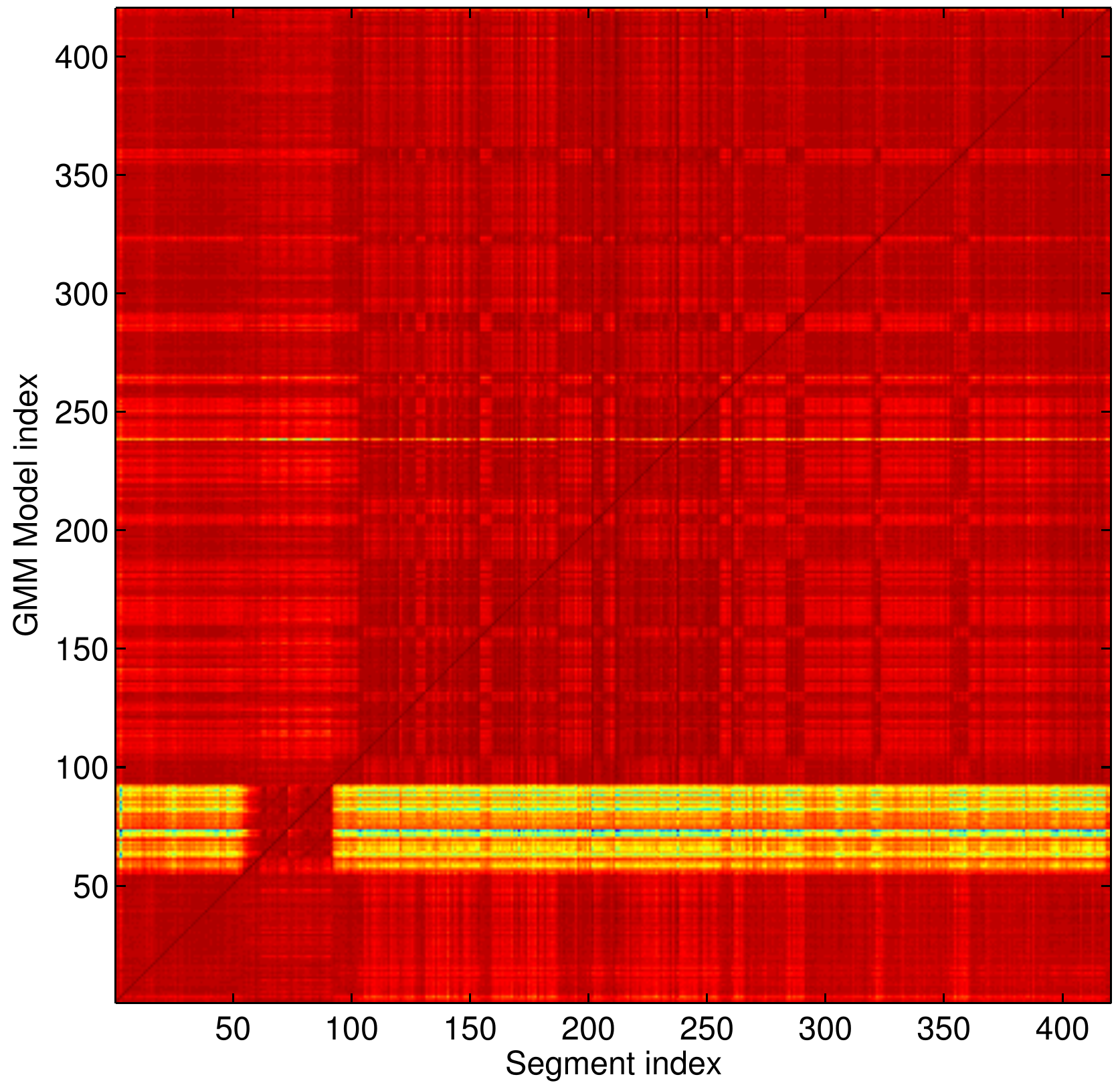}}
            \subfigure[Maximal R vs. \% of outliers\label{fig:maxRscoreAudiohOL}]
            {\includegraphics[width=\factor\linewidth]{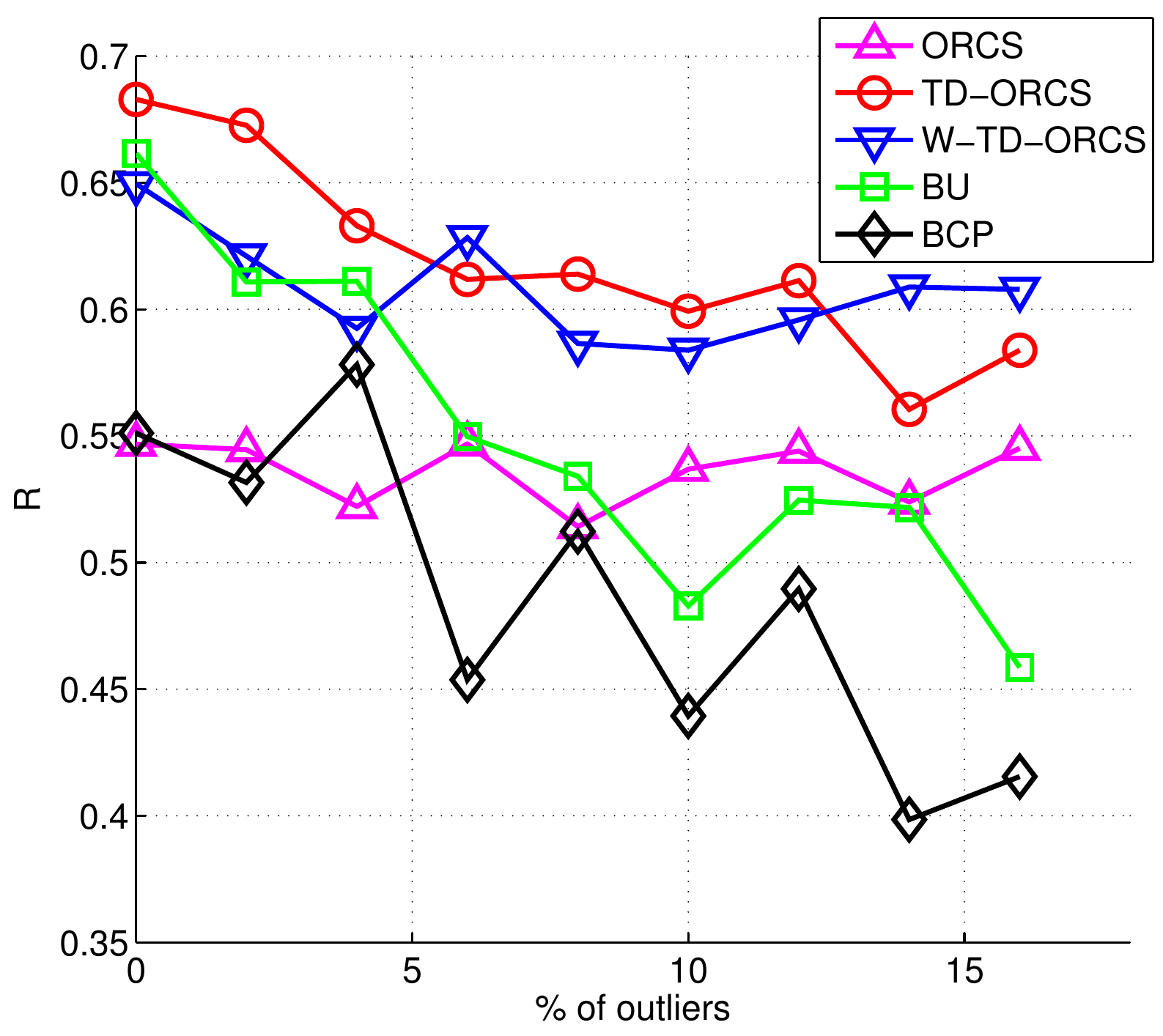}}
            \caption{(a) $A_{ij}$ 
            is the log-probability of segment $j$ given the
            GMM fitted to segment $i$.
            (b) Maximal R measure vs. percentage of
            outliers. See text for algorithms details.
            }
            \centering
            \subfigure[R measure\label{fig:detRateAudio}]{\includegraphics[width=\factor\linewidth]{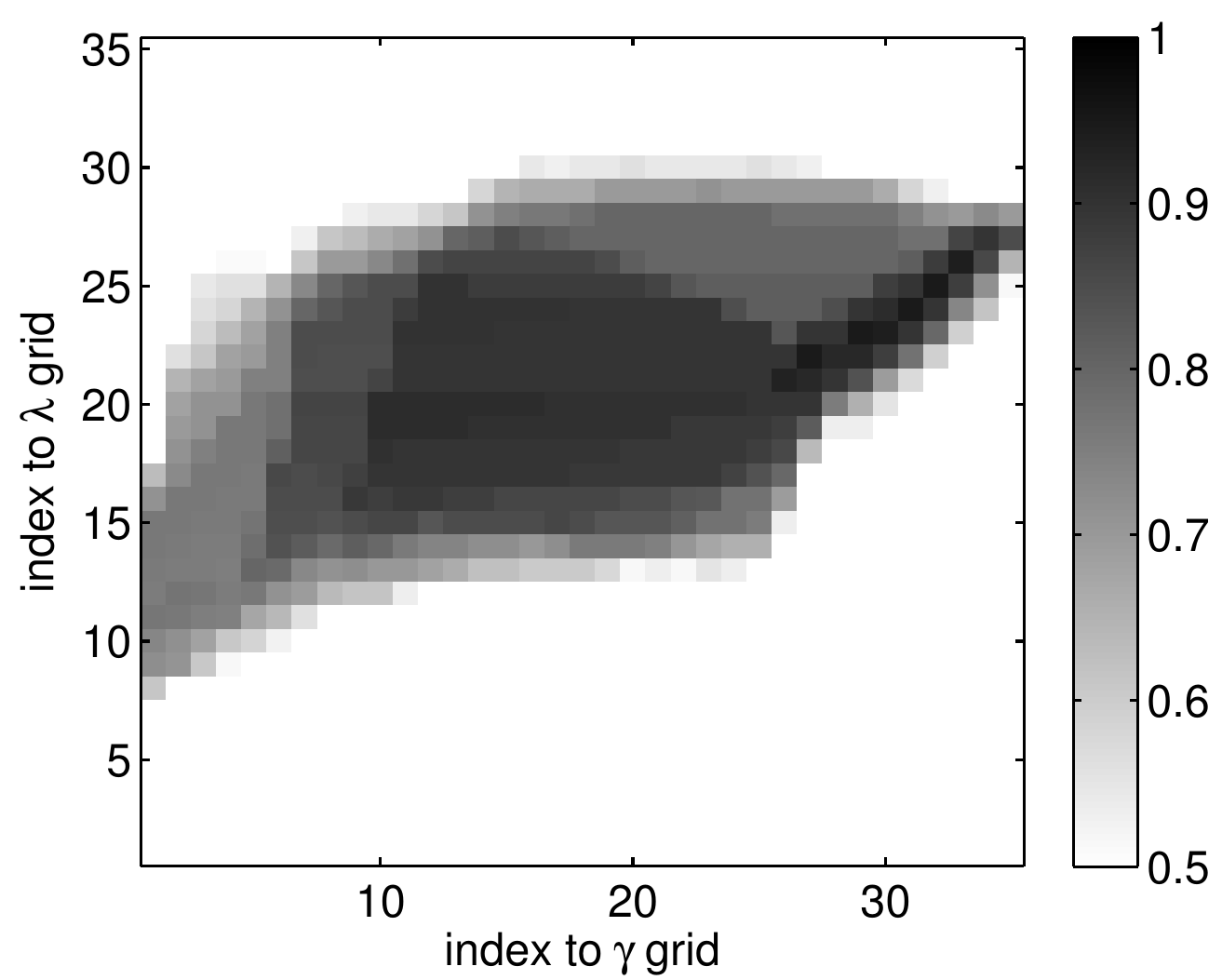}}
            \subfigure[Number of detected outliers\label{fig:nOutliersAudio}]{\includegraphics[width=\factor\linewidth]{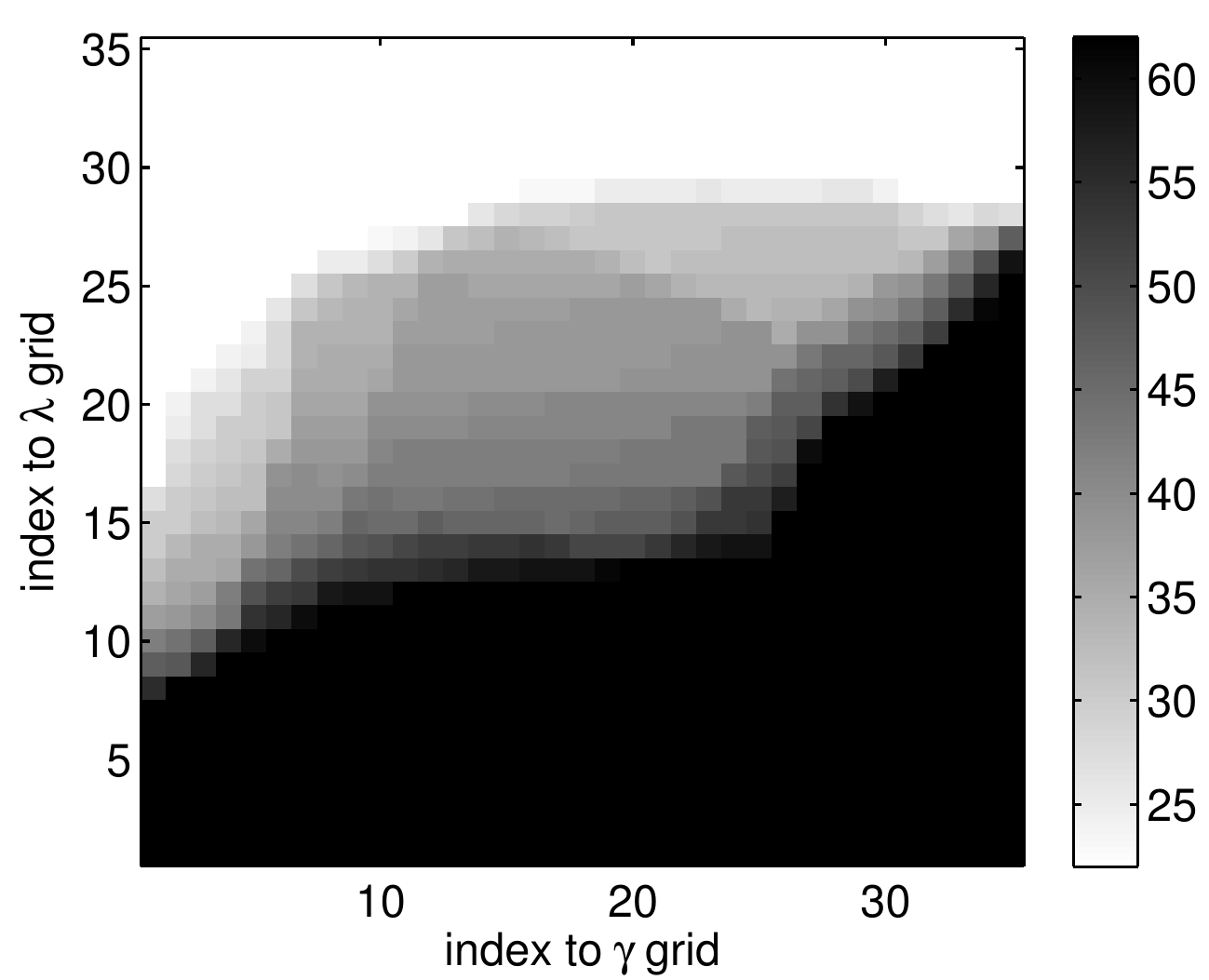}}
            \caption{
            (a) Outliers detection quality by the ORCS algorithm, quantified using the R measure of outliers detection. R is evaluated on a $35\times35$ parameter grid, for $10\%$ outliers.
            (b) The corresponding number of detected outliers.}
            \label{fig:detectionRate}
        \end{minipage}
\end{figure}

\paragraph{Signal representation} A common representation in speech analysis is the MFCC coefficients mentioned in \secref{sec:biphonesSegExperiment}. However, this representation is computed over time windows of tens of milliseconds, and therefore it is not designed to capture the characteristics of a segment with length in the order of seconds or minutes. We therefore apply post-processing on the MFCC representation. First, the raw audio is divided
into $N$ non-overlapping blocks of $5$ seconds duration, and the MFCC coefficients are computed for all blocks $\braces{S_j}_{j=1}^N$. We used 13 MFCC coefficient with $25$ms window length and $10$ms hop length. Then a Gaussian Mixture Model (GMM) $T_j$ with $10$ components and a diagonal covariance matrix is
fitted to the $j$th block $S_j$. These parameters of the GMM were selected using the Bayesian Information Criterion (BIC). The log-likelihood matrix $A$
is then defined by $A_{ij}=\log\prob{S_j|T_i}$. 
The clean feature matrix (no outliers) is shown in
\figref{fig:likemat}, where different segments can be discerned.
Since using the columns of $A$ as features yields a dimension growing 
with $N$, we randomly choose a subset of $d=100$ rows
of $A$, and the columns of the resulting matrix $X\in\mathbb{R}^{d\times N}$ are the
input to the segmentation algorithm.
We repeat the experiment for different number of outliers, ranging between $0\%$ and $16\%$ with intervals of $2\%$.
Outliers were added to the raw audio. A given percentage of outliers refers to the relative
number of blocks randomly selected as outliers, to which we add a $5$ seconds recording of repeated hammer strokes,
normalized to a level of $0$dB SNR.

\paragraph{Algorithms} We consider the Outlier-Robust
Convex Sequential (ORCS) segmentation, and its top-down versions (weighted and unweighted)
which we denote by WTD-ORCS and TD-ORCS, respectively.
We compare the performance to three other algorithms. The first is a greedy bottom-up (BU)
segmentation algorithm, which minimizes the sum of squared errors on
each iteration. The bottom-up approach has been successfully used in
tasks of speech segmentation \cite{qiao:2012study,gracia:2011hierarchical}.
The second algorithm is the W-LARS algorithm of \cite{bleakley2011group}. The third algorithm is a Bayesian change-point detection algorithm (BCP), as formulated by \cite{erdman2008fast}.
A solution path was found as follows.
For the ORCS algorithm, a $35\times35$ parameter grid was used, where
$0<\gamma<\gamma^*$ was sampled uniformly, and for each $\gamma$ value,
$0<\lambda<\lambda^*\paren{\gamma}$ was sampled logarithmically, where
$\lambda^*\paren{\gamma}$ is the critical value for $\lambda$ for a
given choice of $\gamma$ (see \secref{sec:robust_top_down} for details). For the TD-ORCS, W-LARS, and BU algorithms,
$K=2\comdots 150$ number of segments were used as an input to the
algorithms. For the TD-ORCS algorithm, where the number of required
outliers is an additional input parameter, the correct number of
outliers was used. For the BCP algorithm, a range of thresholds on the posterior probability of change-points was used to detect a range of number of segments. As is evident from the empirical results below, the ORCS algorithm can
achieve high detection rate of the outliers even without knowing their
exact number a-priori. Furthermore, we suggest below a way of
estimating the number of outliers.
For each algorithm, the maximal R measure over all parameters range
was used to compare all algorithms.

\paragraph{Results}
Results are shown in
\figref{fig:maxRscoreAudiohOL} as the maximal R measure achieved versus the percentage
of outliers, for each of the algorithms considered. It is evident that the
performance of the BU and BCP algorithms decreases significantly as
more outliers are added, while the outlier-robust ORCS algorithm
keeps an approximately steady performance. Our unweighted and weighted TD-ORCS algorithms
achieve the best performance for all levels of
outliers. Results for LARS algorithm are omitted as it did not perform better than other algorithms.
We verified the ability of our algorithms to correctly detect outliers
by calculating the R measure of the outliers detection of the ORCS algorithm, with zero
length tolerance window, i.e a detection is considered a true-positive
only if it exactly pinpoints an outlier.
The R measure of the detection was evaluated on the $\gamma$, $\lambda$ parameter
grid, as well as the corresponding numbers of detected outliers. Results for the representative case of $p=10\%$ outliers are shown in
\figref{fig:detRateAudio}.
It is evident that a high R measure ($>0.9$) is attained on a range of
parameters that yield around the true number of outliers. We conclude that one does not
need to know the exact number of outliers 
in order to use the ORCS algorithm, and a rough estimate is enough. Some
preliminary results suggest that such an estimate can be approximated
from the histogram of number of detected outliers
(i.e. \figref{fig:nOutliersAudio}).

\section{Related Work and Conclusion}\label{sec:conclusion}
There is a large amount of literature on change-point detection, see for example \cite{basseville1993detection,brodsky1993nonparametric}.
Optimal segmentation can be found using dynamic programming \cite{lavielle2006changepoint}; however, the complexity of this approach is quadratic in the number of samples $n$, and therefore might be infeasible for large data sets. Some approaches which achieve complexity linear in $n$ \cite{levy2007catching,killick2012optimal} treat only one dimensional data. 
Some related work is concerned with the objective \eqref{eq:segmentationObjective_mu} we presented in \secref{sec:algorithms}. In \cite{levy2007catching} it was suggested to reformulate \eqref{eq:segmentationObjective_mu} for the one dimensional case as a LASSO regression problem \cite{tibshirani1996regression,yuan2006model}, while 
\cite{bleakley2011group}
extended this approach to multidimensional data, although not treating outliers directly.
Another common approach is deriving an objective from a maximum likelihood criterion of a generative model, and then either optimize the objective or use it as a criterion for a top-down or a bottom-up approach \cite{qiao:2012study,phonemeSeg:2008,gracia:2011hierarchical,olshen2004circular}. We note that the two-dimensional version of \eqref{eq:segmentationObjective_mu} is used in image denoising applications, where it is known as the Total-Variation of the image \cite{Rudin:1992,Chambolle:2004,Beck_TV:2009}.
Finally, we note that all these approaches do not directly incorporate outliers into the model.


We formulated the task of segmenting sequential data and detecting outliers using convex optimization, which can be solved in an alternating manner. We showed that a specific choice of weighting
can empirically enhance performance.
We described how to calculate
$\lambda^*$ and $\gamma^*$, the critical values for the split into two
segments and the detection of the first outlier, respectively. These
values are useful for finding a solution path in the two-dimensional
parameter space. We also derived a top-down, outlier-robust
hierarchical segmentation algorithm which minimizes the objective in a
greedy manner.
%
This algorithm allows for directly controlling both the
number of desired segments $K$ and number of outliers $M$. Experiments with 
real-world audio data with outliers added manually demonstrated the superiority of our algorithms.

We consider a few possible extensions to the current work. One is deriving algorithms that will work on-the-fly. Another direction is to investigate more involved noise models, such as noise which corrupts a single feature along all samples, or a consecutive set of samples. Yet another interesting question is how to identify that different segments come from the same source, e.g.\ that the same speaker is present at different locations in a recording. We plan to investigate these directions in future work.
%

\bibliography{refs}
\bibliographystyle{aaai}

\clearpage 
\appendix
\section{Proofs}\label{sec:suppMaterial}

\subsection{Proof of \lemref{lem:analyticalSolutionK=2}}\label{sec:proof_lem:analyticalSolutionK=2}
\begin{proof}
Our starting point is the following lemma, which makes further analysis easier.
\begin{lemma}\label{lem:replaceSamplesWithMean}
Assume an optimal solution $\vmu^*$ of \eqref{eq:segmentationObjective_mu} is given, and therefore we also know to which
segment each data sample belongs. If we replace all samples in a segment with the mean of these samples, the optimal
solution $\vmu^*$ will not change.
\end{lemma}
The proof appears in \secref{sec:proof_lem:replaceSamplesWithMean}. 
We now analyze the transition of the solution to \eqref{eq:segmentationObjective_mu} from $K=1$ to $K=2$ segments.
During the analysis we use the fact that the solution path is continuous in $\lambda$, as was shown previously in
another context by \cite{Chi:2013arXiv}. We denote by $\lambda^*$ the value of $\lambda$ at the splitting point, and we
assume that for the two segments solution we have $n_1$ samples in the first segment and $n_2=n-n_1$ samples in the second
segment. We denote the means of the two segments by $\bvxi{1}$ and $\bvxi{2}$. \lemref{lem:replaceSamplesWithMean} allows
us to replace samples in a segment with the mean of the segment, without changing the optimal solution $\vmu^*$.
This means that for $K=2$ all analysis is taking place on the line connecting $\bvxi{1}$ and $\bvxi{2}$ and is
therefore essentially one dimensional. For $K=1$ the regularization term vanishes, so the solution which we denote
by $\vmu$ is simply the mean of the whole data set, $\vmu=\bvxi{1}+\alpha_0(\bvxi{2}-\bvxi{1})$,
where $\alpha_0=n_2/n$. For $K=2$, we denote the solution by $\vmui{1}$ and $\vmui{2}$,
and we parameterize $\vmui{1}$ by $\vmui{1}=\bvxi{1}+\alpha\paren{\bvxi{2}-\bvxi{1}}$, for some $\alpha$.
We note that $\alpha\leq\alpha_0$, since we know that $\vmui{1}$ is closer to $\bvxi{1}$ than $\vmu$ is.
The parametrization for $\vmui{2}$ is therefore $\vmui{2}=\vxi{1}+\paren{1-\frac{n_1}{n_2}\alpha}\paren{\bvxi{2}-\bvxi{1}}$
\footnote{This can be derived either directly by requiring that the derivative of \eqref{eq:segmentationObjective_mu} for $K=2$
equals zeros, or by noting that the `center of mass' of $\vmui{1}$ and $\vmui{2}$ is just $\vmu$. Both approaches gives the
same equation, namely $n_1\vmui{1}+n_2\vmui{2}=n\vmu$.}.
In order to find $\alpha$, we look for a minimum of the objective $h$ for $K=2$:
\begin{align}\label{eq:objK=2}
    h =& \frac{n_1}{2}\normt{\bvxi{1}-\vmui{1}}+\frac{n_2}{2}\normt{\bvxi{2}-\vmui{2}}\\
    &+\lambda w_{n_1}\vnorm{\vmui{2}-\vmui{1}}_2.\nonumber
\end{align}
Plugging $\vmui{1}$ and $\vmui{2}$ as parameterized by $\alpha$ into \eqref{eq:objK=2} and looking for the minimum, we get
\begin{align*}
\alpha & = \argmin{\alpha'}\braces{y^2\alpha'^2{nn_1}/{2n_2}-\lambda w_{n_1} y\alpha'{n}{n_2}+\lambda w_{n_1} y}\\
& = {\lambda w_{n_1}}/{n_1y},\nonumber
\end{align*}
where we define $y\triangleq \vnorm{\bvxi{2}-\bvxi{1}}$. In order to find $\lambda^*$, we require that the objective for $K=1$
and $K=2$ has the same value at the splitting point, where $\lambda=\lambda^*$. This requirement is equivalent to the requirement
that $\alpha=\alpha_0$, since at the splitting point we have $\vmu=\vmui{1}=\vmui{2}$. This leads to the solution for $\lambda^*$,
where we explicitly include the dependence on $n_1$, to emphasize that this is the solution provided that $n_1$ is known:
\begin{equation}\label{eq:lambdaStar_n1}
    \lambda^*(n_1)=\frac{1}{w_{n_1}}\frac{n_1(n-n_1)}{n}\vnorm{\bvxi{2}(n_1)-\bvxi{1}(n_1)}_2.
\end{equation}


In order to find the actual splitting point $n^*$, we note that the split into two segments occurs as $\lambda$ is decreased
from $\lambda>\lambda^*$ to $\lambda<\lambda^*$, so maximizing $\lambda^*(n_1)$ over $n_1$ gives the splitting point $n^*$:
\begin{align*}
    n^*\paren{\vx} &= \argmax{n_1}g\paren{n_1,\vx},\\ 
    \lambda^*\paren{\vx} &= g\paren{n^*,\vx},\nonumber \\
    \textrm{ where: }\\
    g\paren{n_1,\vx} &= \bbraces{\frac{1}{w_{n_1}}\frac{n_1(n-n_1)}{n}\vnorm{\bvxi{2}(n_1)-\bvxi{1}(n_1)}_2}.\nonumber
\end{align*}
\end{proof}\QED
\subsection{Proof of \lemref{lem:weighted_solution_l0_l1}}\label{sec:proof_lem:weighted_solution_l0_l1}
\begin{proof}
For $K=2$ the unrelaxed optimization problem \eqref{eq:segmentationObjective_unrelaxed_K=2} is equivalent to
\begin{align*}
    \argmin{n_1,\vmui{1},\vmui{2}}\braces{\sum_{i=1}^{n_1}\normt{\vxii-\vmui{1}}+\sum_{i=n_1+1}^{n}\normt{\vxii-\vmui{2}}}~.
\end{align*}
The minimization on $\vmui{1},\vmui{2}$ is immediate and is given by the means of the segments, so we get
\begin{align}\label{eq:2seg_l0constraint}
    \argmin{n_1}\braces{\sum_{i=1}^{n_1}\normt{\vxii-\bar{x}_1}+\sum_{i=n_1+1}^{n}\normt{\vxii-\bar{x}_2}}~,
\end{align}
where we defined
\begin{align*}
\bar{x}_1&=\frac{1}{n_1}\sum_{i=1}^{n_1}\vxii\\
\bar{x}_2&=\frac{1}{n_2}\sum_{i=n_1+1}^{n}\vxii~
\end{align*}
and $n_2\triangleq n-n_1$.
Recall that the solution to the relaxed optimization problem is given by \lemref{lem:analyticalSolutionK=2}, as described in \secref{sec:outlierRobustObj}:
\begin{align}\label{eq:2seg_weighted_l1}
    \argmax{n_1,n_2=n-n_1}\braces{\frac{n_1 n_2}{nw_{n_1}}\vnorm{\bar{x}_2-\bar{x}_1}_2},
\end{align}
where $w_i$ are the weights. We now argue that \eqref{eq:2seg_l0constraint} and \eqref{eq:2seg_weighted_l1} have the same solution, for the specific choice of $w_i=\sqrt{i\paren{n-1}/n}$. First note that \eqref{eq:2seg_l0constraint} can be rewritten as $\argmax{n_1,n_2=n-n_1}\braces{n_1\normt{\bar{x}_1}+n_2\normt{\bar{x}_2}}$. Next, we show that this objective and the square of the (non-negative) objective \eqref{eq:2seg_weighted_l1} differ by a constant $C$, which depends on the data $\vxii$ but not on $n_1$:
\begin{align*}
    n_1\normt{\bar{x}_1}+n_2\normt{\bar{x}_2}&=\frac{n_1 n_2}{n}\normt{\bar{x}_2-\bar{x}_1}+C\\
    \frac{n_1^2}{n}\normt{\bar{x}_1}+\frac{n_2^2}{n}\normt{\bar{x}_2}&=C-\frac{2n_1 n_2}{n}\bar{x}_1^T\bar{x}_2\\
    \frac{1}{n}\normt{\sum_{i=1}^n\vxii}&=C,
\end{align*}
which proves that indeed  \eqref{eq:2seg_l0constraint} and \eqref{eq:2seg_weighted_l1} attain their optimal value at the same $n_1^*$
\end{proof}\QED
\newpage
\subsection{Proof of \eqref{eq:1D_errBound_m}}\label{sec:proof:1D_errBound_m}
\begin{proof}
Define the random variable
$Y_m$ which is the difference between the empirical means of two
subsequences created by splitting after $n_1+m$ samples:
\begin{align*}
    Y_m &\triangleq\frac{1}{n_2-m}\negspaces\sum_{i=n_1+m+1}^{n}\negspace\vxii-\frac{1}{n_1+m}\sum_{i=1}^{n_1+m}\vxii.
\end{align*}
This allows us to rewrite
\eqref{eq:lambdaStarWeighted} as an optimization over $m$:
\begin{align*}
    g\paren{m} &\triangleq \frac{(n_1+m)(n_2-m)}{n}\abs{Y_m},\\
    m^* &=\argmax{m}\braces{g\paren{m}},
\end{align*}
where $m\in\brackets{-n_1+1,n_2-1}$.
Without loss of generality, we treat the case where
$m\geq0$. Note that $m^*\neq0$ if $g(0)<g(m)$ for some $m>0$.
The probability of this event is:
\begin{align}\label{eq:correctBoundaryErrProb}
    &\prob{g(0) < g(m)}=\\
    &\prob{\frac{n_1n_2}{n}\abs{Y_0}<\frac{(n_1+m)(n_2-m)}{n}\abs{Y_m}}.\nonumber
\end{align}
Defining the following random variables:
\begin{equation}\label{eq:Wdef}
    W_m^{\pm}\triangleq\frac{n_1n_2}{n}Y_0\pm\frac{(n_1+m)(n_2-m)}{n}Y_m,
\end{equation}
we can rewrite
\begin{align}\label{eq:errBound_asW}
    \prob{g(0)<g(m)} & \leq\prob{W_m^{+}<0}+\prob{W_m^{-}<0},
\end{align}
where we used the fact that
for two random variables $A$ and $B$, it holds true that
\begin{align*}
    \prob{\abs{A}<\abs{B}}\leq\prob{A<B}+\prob{A<-B}.
\end{align*}
Using the definition of $Y_m$, we rewrite $W_m^{\pm}$
as the (weighted) average of a sequence of the $n$ random variables composing the data,
\begin{align}
    W_m^+=\frac{1}{n}\Bigg(&\paren{m-2n_2}\sum_{i=1}^{n_1}\vxii+\paren{n_1-n_2+m}\negspaces\sum_{i=n_1+1}^{n_1+m}\vxii\nonumber\\
    &+\paren{2n_1+m}\negspace\sum_{i=n_1+m+1}^{n}\negspace\vxii\Bigg),\nonumber
\end{align}
and
\[
    W_m^-=\frac{1}{n}\paren{-m\sum_{i=1}^{n_1}\vxii+\paren{n-m}\negspaces\sum_{i=n_1+1}^{n_1+m}\negspaces\vxii-m\negspace
    \sum_{i=n_1+m+1}^{n}\negspace\vxii}.
\]
The means of $W_m^{\pm}$ are given by
\begin{align*}
    \expec{W_m^-} &= \frac{n_1m}{n}\Delta\mu,\\
    \expec{W_m^+} &= \frac{n_1\paren{2n_2-m}}{n}\Delta\mu,
\end{align*}
where we define $\Delta\mu\triangleq\mu_2-\mu_1$. From \eqref{eq:correctBoundaryErrProb}, \eqref{eq:Wdef}, and \eqref{eq:errBound_asW} it follows that
\begin{align*}
    &\prob{\frac{n_1n_2}{n}Y_0<\frac{(n_1+m)(n_2-m)}{n}\abs{Y_m}}\\
    \leq&\prob{W_m^-<0}+\prob{W_m^+<0}.\nonumber
\end{align*}
Using Hoeffding's inequality to bound the probabilities that $W_m^{\pm}$ are negative we get:
\begin{align*}
    \prob{W_m^-<0}\leq\exp\paren{-\frac{2n_1^2m\Delta\mu^2}{M^2n\paren{n-m}}}\triangleq B^-,
\end{align*}
and similarly:
\begin{align*}
    &\prob{W_m^+<0}\\
    &\leq\exp\paren{-\frac{2n_1^2\paren{2n_2-m}^2\Delta\mu^2}{M^2n\paren{m\paren{n-m-4n_1}+4n_1\paren{n-n_1}}}}\\
    &\triangleq B^+.
\end{align*}
It can be shown that $B^+<B^-$, provided that $m$ is in its feasible range, i.e $0<m<n_2$. We conclude that
\begin{align*}
    &\mathbb{P}\paren{\frac{n_1n_2}{n}\abs{Y_0}<\frac{(n_1+m)(n_2-m)}{n}\abs{Y_m}}\\
    &\leq B^-+B^+\leq 2B^-,
\end{align*}
so we have that
\begin{align*}
    &\mathbb{P}\paren{\frac{n_1n_2}{n}\abs{Y_0}<\frac{(n_1+m)(n_2-m)}{n}\abs{Y_m}}\\
    &\leq2\exp\paren{-\frac{2n_1^2m\Delta\mu^2}{M^2n\paren{n-m}}}\leq2\exp\paren{-\frac{2n_1^2\Delta\mu^2}{M^2n^2}m},
\end{align*}
which proves \eqref{eq:1D_errBound_m} for
$C=\frac{2\Delta\mu^2n_1^2}{M^2n^2}$.
\end{proof} \QED
\newpage
\subsection{Proof of \thmref{thm:1D_errBound}}\label{sec:proof:1D_errBound}
\begin{proof}
Using \eqref{eq:1D_errBound_m} and the union bound, we get
\begin{align*}
    \prob{m^*\geq m_0} &\leq \prob{\exists m\geq m_0:\:g(0)<g(m)} \\
    &\leq\sum_{m=m_0}^{n_2}\prob{g(0)<g(m)} \\
    &\leq n_2\prob{g\paren{0}<g\paren{m_0}} \\
    &\leq 2n_2\exp\paren{-Cm_0} \stackrel{(\star)}{\leq} \delta~,
\end{align*}
where ($\star$) follows from 
the definition of $m_0$.
\end{proof}\QED

\subsection{Proof of \lemref{lem:replaceSamplesWithMean}}\label{sec:proof_lem:replaceSamplesWithMean}
\begin{proof} Consider a segment of $n_0$ data samples, and denote by $\vmui{0}$ the centroid of this segment. The mean
of the segment is given by $\bvxi{0}=\frac{1}{n_0}\sum\limits_{\vxii\in\vmui{0}}\vxii$. The contribution of this segment to
the first term in the objective \eqref{eq:segmentationObjective_mu} is given by
\begin{equation}\label{eq:oneSegContribution}
\frac{1}{2}\sum\limits_{\vxii\in\vmui{0}}\negspaces\normt{\vxii-\vmui{0}}=
\frac{1}{2}\sum\limits_{\vxii\in\vmui{0}}\negspaces\left(\normt{\vxii}-2\vxii^T\vmui{0}+\normt{\vmu_0}\right).
\end{equation}
If we now substitute $\bvxi{0}$ for each sample $\vxii$ in this segment, the contribution to the objective becomes
\begin{equation}\label{eq:oneSegMeanContribution}
    \frac{n_0}{2}\normt{\bvxi{0}-\vmui{0}}=\frac{n_0}{2}\left(\normt{\bvxi{0}}-2\bvxi{0}^T\vmui{0}+\normt{\vmui{0}}\right).
\end{equation}
It is straightforward to show that as a function of $\vmui{0}$, \eqref{eq:oneSegContribution} and
\eqref{eq:oneSegMeanContribution} differ by a constant which depends only on the data samples $\vxii\in\vmui{0}$,
not on $\vmui{0}$. Since this argument holds for all segments, we conclude that replacing each data sample with the mean
of the segment to which it belongs, results in the same objective, up to a constant. Therefore the optimal solution $\vmu^*$
does not change.
\end{proof}\QED

\end{document}